\documentclass{article}

\usepackage{microtype}
\usepackage{graphicx}
\usepackage{subfigure}
\usepackage{booktabs} % for professional tables
\usepackage{multirow}
\usepackage{adjustbox}
\usepackage{pifont}
\usepackage{amssymb}
\usepackage{xcolor}
\usepackage{float}
\usepackage{listings}

\usepackage{hyperref}

\newcommand{\cmark}{\ding{51}}%
\newcommand{\xmark}{\ding{55}}%

\usepackage[accepted]{icml2019}

\icmltitlerunning{Semi-Conditional Normalizing Flows for Semi-Supervised Learning}

\usepackage[normalem]{ulem} 

\usepackage{hyperref}
\usepackage{url}
\usepackage{amsmath}
\usepackage{amsmath}
\usepackage{amsfonts}
\usepackage{graphicx}
\usepackage{multirow}
\usepackage{wrapfig}
\usepackage{xfrac}

\usepackage{booktabs}
\usepackage{blindtext}
\usepackage{graphicx}
\usepackage{etoolbox}
\usepackage{bm}
\usepackage{scrextend}
\usepackage[flushleft]{threeparttable}
\usepackage{amsmath,amsthm, amssymb, latexsym}
\usepackage{epsfig}
\usepackage{multirow}
\usepackage{pifont}
\usepackage{bigints}
\usepackage{soul}
\usepackage{marginnote}
\usepackage{colortbl}
\usepackage{xcolor}

\DeclareMathOperator{\N}{\mathcal{N}}

\DeclareMathOperator{\R}{\mathbb{R}}

\begin{document}

\twocolumn[
\icmltitle{Semi-Conditional Normalizing Flows for Semi-Supervised Learning}



\icmlsetsymbol{equal}{*}

\begin{icmlauthorlist}
\icmlauthor{Andrei Atanov}{hse,skoltech}
\icmlauthor{Alexandra Volokhova}{mipt,ysda}
\icmlauthor{Arsenii Ashukha}{saic}
\icmlauthor{Ivan Sosnovik}{bosch}
\icmlauthor{Dmitry Vetrov}{saic,hse}
\end{icmlauthorlist}

\icmlaffiliation{hse}{Samsung-HSE Laboratory, National Research University Higher School of Economics}
\icmlaffiliation{skoltech}{Skolkovo Institute of Science and Technology}
\icmlaffiliation{saic}{Samsung AI Center Moscow}
\icmlaffiliation{mipt}{Moscow Institute of Physics and Technology}
\icmlaffiliation{bosch}{UvA-Bosch Delta Lab, University of Amsterdam, Netherlands}
\icmlaffiliation{ysda}{Yandex School of Data Analysis}

\icmlcorrespondingauthor{Andrei Atanov}{ai.atanow@gmail.com}
\icmlcorrespondingauthor{Alexandra Volokhova}{s-volohova@yandex.ru}
\icmlcorrespondingauthor{Arsenii Ashukha}{ars.ashuha@gmail.com}
\icmlcorrespondingauthor{Ivan Sosnovik}{sosnovikivan@gmail.com}

\icmlkeywords{Machine Learning, ICML}

\vskip 0.3in
]



\printAffiliationsAndNotice{\icmlEqualContribution} % otherwise use the standard text.

\begin{abstract}
% Semi-supervised learning approaches based on deep generative models show strong empirical results [Kingma].
% They benefit from using informative low-dimension representation of data provided by a generative model.
% Normalizing Flows (NFs) framework allows for fitting a complex parametric distribution directly, without using a variational approach.
% However, NFs don't provide a direct way for dimension reduction.
% In this paper we utilise multi-scale flow architecture to learn informative low-dimensional representation.
% While NFs cannot deal with discrete variables, we model conditional distribution over objects given label, rather then joint one, by conditioning flow transformations.
% To model this distribution over latent representations rather then raw data we proposed semi-conditional NFs, where only a small number of transformations is conditioning on a label.
% We demonstrate the performance of our approach using modern Glow and NeuralODE based flows on MNIST semi-supervised benchmark.

% Semi-supervised learning is an important field of machine learning research. Deep generative models is known to be helpful in this task.
% We introduce an explicit semi-conditional generative model which efficiently uses labelled and unlabelled data for semi-supervised classification task. Our model allows directly maximize log-likelihood of both labeled and unlabeled data and straightforward estimate posterior distribution over labels for making prediction. We explored efficiency of our model experimentally and found that it outperforms baseline approach.

This paper proposes a semi-conditional normalizing flow model for semi-supervised~learning.
The model uses both labelled and unlabeled data to learn an explicit model of joint distribution over objects and labels.
Semi-conditional architecture of the model allows us to efficiently compute a value and gradients of the marginal likelihood for unlabeled objects. 
The conditional part of the model is based on a proposed \textit{conditional coupling layer}.
We demonstrate a performance of the model for semi-supervised classification problem on different datasets.
The model outperforms the baseline approach based on variational auto-encoders on MNIST dataset. 
% The model outperforms the baseline approach based on variational auto-encoders and achieves $1.9\%$ average error on MNIST dataset with 100 labeled examples. 

\end{abstract}
\vspace{5pt}
\section{Introduction}
\label{submission}
Modern supervised machine learning approaches require large amount of labelled data.
However, large labelled datasets are not always available and cannot be collected without additional effort.
In many fields, on the other hand, unlabelled data is accessible.
This makes semi-supervised learning an important field of machine learning research.
% Modern supervised machine learning approaches require a large amount of labelled data.
% Labelling new data, however, is usually difficult and expensive task.
% On the other hand, large amount of unlabelled data is available in many fields.
% This makes semi-supervised learning an important practical problem.
% In many fields only a limited amount of labelled data is available.
% While a huge amount of unlabelled data is available, labelling them is usually a challenging and expensive task.
Semi-supervised methods aim to employ unlabelled data to achieve a competitive performance of supervised models while using only a few labelled examples.
% Semi-supervised methods are able to effectively employ unlabelled data with only a few labelled examples.
% The long line of research has been done on the topic, it includes but not limited to
Previous work on this topic includes
\citet{joachims1999transductive,blum2004semi,rosenberg2005semi}.
% A number of methods extending classical supervised approaches have been proposed so far \cite{joachims1999transductive, rosenberg2005semi}.
% These methods, however, have difficulties in extending to high dimensions and large amount of unlabelled data.
% These methods, however, are not suitable for high dimensional data and poorly scale on huge unlabelled datasets \citet{zhu2005semi, kingma2014semi}.
These methods, however, poorly scale on huge unlabelled datasets and high dimensional data \citet{zhu2005semi, kingma2014semi}.
% A more detailed related work can be found in \citet{zhu2005semi, kingma2014semi}.

% Recent advances in deep generative models allow to capture complex data structure in unsupervised fashion \cite{kingma2013auto,goodfellow2014generative,dinh2014nice,van2016conditional}.
% Semi-supervised learning methods based on deep generative models outperform previous approaches, especially on image data \cite{springenberg2015unsupervised,kingma2014semi,salimans2016improved}.
% Recent advances in unsupervised learning \cite{kingma2013auto,goodfellow2014generative,dinh2014nice,van2016conditional} allow to model complex data distribution $p(x)$ by using deep generative models.
Recent advances in unsupervised learning, specifically deep generative models \cite{kingma2013auto,goodfellow2014generative,dinh2014nice,van2016conditional}, allow to model complex data distribution $p_\theta(x)$.
Modern methods for semi-supervised learning \cite{kingma2014semi,springenberg2015unsupervised,salimans2016improved} actively employ deep generative models to learn from unlabeled data.
The common approach suggests to model \textit{joint} distribution $p_\theta(x, y) = p_\theta(x|y)p(y)$ over objects $x$ and labels $y$, where $p_\theta(x|y)$ is a conditional generative model.
The joint distribution allows inferring class-label distribution $p_\theta(y|x)$ that is used to make a prediction for a supervised problem.

\citet{kingma2014semi} proposed a semi-supervised learning method that is based on variational auto-encoders \cite{kingma2013auto}.
This approach scales well on large datasets and provides flexible generative models for high-dimensional data.
However, direct likelihood optimization in this case is intractable, and it is usually tackled by methods for approximate variational inference \cite{paisley2012variational, hoffman2013stochastic, ranganath2014black}.

Recently a new family of deep generative models called normalizing flows has been proposed \cite{rezende2015variational}.
Normalizing flows model data by a deterministic invertible transformation of a simple random variable, e.g. standard normal.
The framework provides a tractable likelihood function and allows its direct optimization.
Data distribution in this case can be inferred using change of variables formula by computing Jacobian determinant of the transformation.
Recent advances in this field  \cite{kingma2016improved, dinh2016density, kingma2018glow, grathwohl2018ffjord, berg2018sylvester} provide a number of flexible architectures with tractable and efficient determinant computation.

It has been shown that dimension reduction is crucial for semi-supervised methods \cite{kingma2014semi}.
Unfortunately, normalizing flows have a latent representations of the same dimension as input data.
However, it was shown that a multi-scale architecture \cite{dinh2016density} produces latent representation, in which only a small part of components stores consistent semantic information about input data.
That opens a possibility to use this kind of architecture for dimension reduction.

\vspace{.5em}
In this paper, we propose \textit{Semi-Conditional Normalizing Flow} --- a new normalizing flow model that is suitable for semi-supervised classification problem.
The proposed model provides a tractable likelihood estimation for both labeled and unlabeled data. 
The semi-conditional architecture (Section~\ref{sec:SCNF-model}) allows us to effectively compute value and gradients of a marginal likelihood  $p_\theta(x)$.
The conditional distribution $p_\theta(x|y)$ is modelled by a proposed \textit{conditional coupling layer}, that is built on the top of the original coupling layer \cite{dinh2016density}.
For dimension reduction, we adapt a multi-scale architecture, and we find it to be a crucial component of our model.
In experiments, we demonstrate the empirical performance of the model for semi-supervised classification problem.
We also find that the proposed model incorporates a data obfuscation mechanism by design, that may be useful for semi-supervised fair learning. 

\vspace{-5pt}

\section{Semi-Supervised Learning with \newline Deep Generative Models}
\label{sec:ssl}

We consider semi-supervised classification problem with partly labelled dataset. 
Let us denote an object as $x_i \in \R^d$ and a corresponding class label as $y_i \in \{1, \dots, K\}$. We denote the set of labelled and unlabelled objects as $\mathcal{L}$ and $\mathcal{U}$ respectively. 
Generative model with parameters $\theta$ defines joint likelihood $p_{\theta}(x, y)$ of an object and its label.
The objective function for this model is a log-likelihood of the dataset: 
\begin{equation}
    \label{eq:ssl-objective}
    {L}(\theta) = \sum_{(x_i, y_i) \in \, \mathcal{L}}{\!\!\!\!\log{ p_{\theta}(x_i, y_i)}} + \sum_{x_j \in \, \mathcal{U}}{\log{p_{\theta}(x_j)}},
\end{equation}
where $p_{\theta}(x_j) = \sum_{k=1}^K p_{\theta}(x_j, y\!\!=\!\!k)$ is a marginal likelihood for unlabelled object. 
The joint likelihood $p_{\theta}(x, y)$ is often parameterized as $p_{\theta}(x, y) = p_{\theta}(x \vert y) p(y)$,
where a prior $p(y)$ is a uniform categorical distribution and a conditional distribution $p_{\theta}(x \vert y)$ is defined by a conditional generative model.
On the test stage, the prediction of the model -- a posterior distribution over labels $p_{\theta}(y \vert x)$ -- can be found as $p_{\theta}(y \vert x) = \tfrac{p_\theta(x, y)}{p_{\theta}(x)}$. 

\citet{kingma2014semi} proposed to use variational auto-encoders \cite{kingma2013auto} as the conditional generative model $p_{\theta}(x \vert y)$.
However, such parametrization, does not allow to maximize the objective \eqref{eq:ssl-objective} directly and obliges to use stochastic optimization of a variational lower bound, that does not guarantee convergence to the maximum of  \eqref{eq:ssl-objective}. 
Additionally, the posterior predictive distribution $p_{\theta}(y \vert x)$  cannot be computed in a closed form and needs a time-consuming sample-based estimation.
% Additionally, the posterior predictive distribution $p_{\theta}(y \vert x)$ is not tractable.
% In this case, the conditional distribution $p_{\theta}(x \vert y)$ equals to $\int p_{\theta}(x|z,y) p(z) dz$. Here $z$ is a continuous latent variable with standard normal prior $p(z)$ and $p(x|z,y,\theta)$ is a simple parametric distribution (e.g. Gaussian), which parameters are modelled by a Deep Neural Network.

We present a new way to model the conditional distribution $p_{\theta}(x\vert y)$ using normalizing flows \cite{rezende2015variational}.
In our approach we have access to all required distributions to maximize log-likelihood of the data
\eqref{eq:ssl-objective} directly and to compute exact posterior $p_{\theta}(y \vert x)$ see (Section~\ref{sec:SCNF-model}).

\section{Semi-Conditional Normalizing Flows}
\label{sec:SCNF}
\subsection{Normalizing Flows}
Normalizing flows model data as a deterministic and invertible transformation $x = g_\theta(z)$ of a simple random variable $z$, e.g. standard normal.
We denote an inverse transformation $g^{-1}_\theta$ as $f_\theta$, i.e. this inverse function maps data $x$ to latent representation $z$.
% Throughout the paper we will use an inverse notation $z = f_\theta(x)$, where $f_\theta=g_{\theta}^{-1}$.
% Distribution of data given by this procedure can be then calculated using change of variables formula:
Log-density of this model can be calculated using change of variable formula:
\begin{equation}
    \label{eq:nf-logp}
    \log p_\theta(x) = \log \left| \dfrac{\partial f_\theta(x)}{\partial x^T} \right| + \log p(z),
\end{equation}
where $\frac{\partial f_\theta(x)}{\partial x^T}$ is a Jacobian of the transformation $f_\theta$ and prior $p(z)$ is a standard normal distribution $\N(z \vert 0, I)$. 
The transformation $f_\theta$ has to be bijective and has a tractable Jacobian determinant.
% A number of such bijective transformations with simple Jacobian matrix have been proposed recently \cite{kingma2016improved, dinh2016density, kingma2018glow}.
% Stacking such simple blocks one can build a very flexible transformation and the corresponding probability is given by a chain rule.
A number of flow architectures \cite{dinh2014nice, kingma2016improved, kingma2018glow} have been proposed recently.
These architectures are based on operations with simple form determinant of Jacobian.
For example, Real NVP architecture proposed by \citet{dinh2016density} is based on an affine coupling layer.
This layer splits an input variable $x$ into two non-overlapping parts $x_1, x_2$ and applies affine transformation based on the first part $x_1$ to the other $x_2$ in the following way:
\begin{equation*}
    z_1 = x_1, ~~~ z_2 = x_2 \odot \exp(s(x_1)) + t(x_1),
\end{equation*}
where $s, t$ are arbitrary neural networks.
This transformation has a simple triangular Jacobian, and the use of neural networks allows to model complex non-linear dependencies.

\subsection{Dimension Reduction with Normalizing Flows}
\label{sec:dim}
It has been shown that performance of a semi-supervised model benefits from using low-dimensional data representation \cite{kingma2014semi}.
% (we also observe this in our experiments, Section~\ref{sec:low-dim}).
% Actually, it is very natural as complex models prune to overfitting in high-dimensional  spaces with few training examples [cite?].
% This can be caused by an overfitting problem as we have only a few labelled objects high-dimension space.
We also observe this effect in our experiments (Section~\ref{sec:exp-dim}).
% Unfortunately, normalizing flows have no such low-dimension representations and preserve a dimension of $z$ equals to one of $x$.
% Unfortunately, normalizing flows preserve a dimension of $z$ equals to one of $x$.
Unfortunately, normalizing flows preserve a dimensionality of $z$ equal to $x$ one.
% However, modern architectures for image generation have multiscale structure \cite{dinh2016density}, where only a part of the variable $x$ pass all transformations.
% However, modern normalizing flow models have multiscale structure, where differnet parts of the variable $x$ pass different number of transformations.
However, modern normalizing flow models have multiscale structure, where differnet parts of the latent representation $z$ pass different number of transformations.
% The deepest part that passes all transformations contains semantic information about an input object \cite{dinh2016density},
% \textcolor{blue}{we will refer to this part as $z_f$ and to the other components as $z_\mathrm{aux}$.}
The part that passes all transformations contains most of semantic information about an input object \cite{dinh2016density},
we will refer to this  part as $z_f$ and to the other components as $z_\mathrm{aux}$.

% Thus, we pass only the deepest components of $z$ (we refer to them as $z_f$) into conditional part and the other components $z_\mathrm{aux}$ directly to standard normal distribution (Fig.~\ref{fig:scnf-model}).

\subsection{Semi-Conditional Normalizing Flow}
\label{sec:SCNF-model}
%\textbf{Conditional Normalizing Flows}
Semi-supervised methods based on generative models require conditional density estimation (Section~\ref{sec:ssl}).
In order to model conditional distribution $p_\theta(x\vert y)$ rather then just $p_\theta(x)$ we have to condition the transformation $f_\theta$ on $y$.
% This new transformation still has to be invertible and provide tractable Jacobian wrt $x$.
% Perhaps, the simplest way will be to predict different parameters for each class \cite{}.
% This, however, leads to high memory cost and doesn't provide weight sharing for different classes.
Perhaps, the simplest way would be to use different flows for each class \cite{trippe2018conditional}.
Nevertheless, this approach has high memory cost and does not share weights between classes.
In \cite{kingma2016improved} authors proposed a memory efficient conditional autoregressive architecture.
Unfortunately, sequential structure of autoregressive models leads to high computational cost.
Therefore, we adapt a more memory and computationally efficient affine coupling layer and propose a \textit{conditional affine coupling layer} defined as follows:
\begin{equation*}
    z_1 = x_1, ~~~ z_2 = x_2 \odot \exp(s(x_1, y)) + t(x_1, y),
\end{equation*}
where neural networks $s, t$ have a class variable as an additional input. 
This allows to model complex dependencies on the class variable and at the same time keeps determinant of Jacobian easy to compute.

\begin{figure}[]
    \centering
    \includegraphics[width=0.48\textwidth]{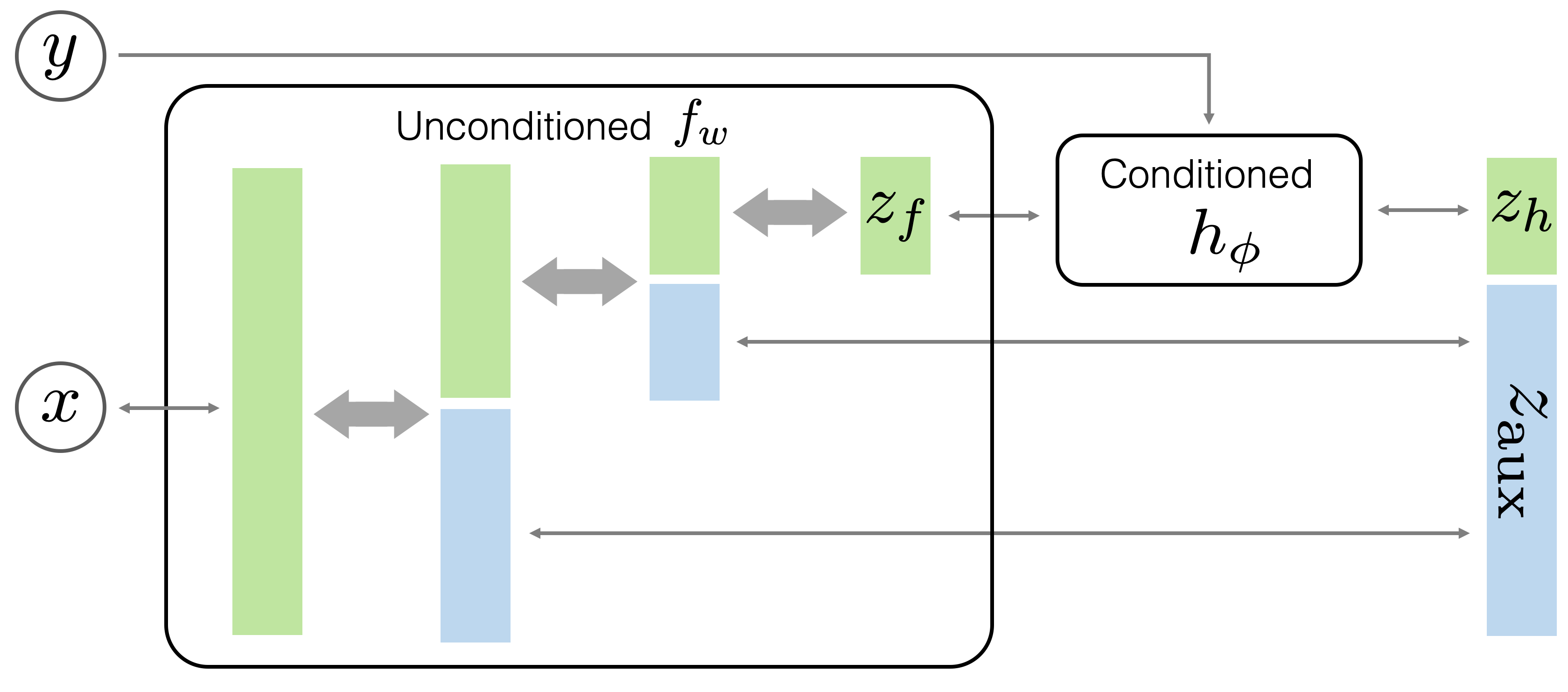}
    % \caption{Semi-Conditional Normalizing Flow model. Input vector $x$ sequentially passes trough invertible transformations (big grey arrows) followed by splitting in two parts. The first pa}
    \caption{
    % Semi-Conditional Normalizing Flow model. Our model consists of unconditioned flow $f_{\theta}$ and conditioned flow $h_{\phi}$. 
    % Unconditioned flow takes object $x$ as an input and gives $z_f$ and $z_{aux}$ as an output.
    The proposed semi-conditional architecture consists of two parts: a large unconditional flow $f_w(x)$, and a relatively small conditional flow $h_\phi(z_f; y)$.
    The unconditional flow $f_w(x)$ is based on a multi-scale architecture and maps an input $x$ into a low-dimensional $z_f$ and an auxiliary vector $z_\mathrm{aux}$.
    The conditional flow $h_\phi(z_f; y)$ maps the low-dimensional vector $z_f$ to $z_h\!=\!h_\phi(z_f; y)$.
    % The vector $z_f$ is then passes through the conditional flow $h_\phi$ into $z_h$.
    The architecture allows to compute $p_\theta(x)\!=\!\mathbb{E}_y p_\theta(x, y)$ with a single forward pass of the computationally expensive flow $f_\theta$ and one pass of the inexpensive flow $h_\phi$ for every class label $y$.
    % \vspace{-1em}
    }
    \label{fig:scnf-model}
\end{figure}

\textbf{Semi-Conditional Architecture} 
% True posterior over class variable \eqref{eq:predictive-posterior} is required for prediction and during optimisation algorithm (see Section~\ref{sec:SCNF-opt}).
During the optimisation of the objective \eqref{eq:ssl-objective}, we need to evaluate marginal log-likelihood for unlabelled data $p_\theta(x) = \sum_{y=1}^K  p_\theta(x|y) p(y)$. 
However, this requires $K$ forward passes.
% It computation requires marginalisation over $y$ which has $K$-times higher computational cost as we have to perform forward pass for all possible values of $y$.
We address this issue with a proposed \textit{semi-conditional} architecture where only a small number of deep layers are conditioned on $y$.
First, we map $x$ to $z_f$ and $z_\mathrm{aux}$ with an unconditional flow $f_w(x)$ and then map the deepest components $z_f$ to $z_h$ with a conditional $h_\phi(z_f; y)$.
In this case, the Jacobian of the unconditional flow $f_w$ is independent of $y$, and we can pull it out of the sum and compute only once for all classes:
\begin{align}
\label{eq:scnf-mll}
    % \scriptstyle
   &~~\log p_\theta(x) = \log \left| \tfrac{\partial f_w(x)}{\partial x^T} \right| + \log \N(z_{\mathrm{aux}}|0, I)  \\
    \nonumber&+ \log \left( \sum_{y=1}^K \left| \tfrac{\partial  h_\phi(z_f; y)}{\partial z_f^T} \right| \cdot \N(z_h|0, I)  \cdot p(y) \right)
\end{align}
% \begin{align}
% \label{eq:mll}
%     p_\theta(x)
%     % \nonumber
%   &=  \left| \dfrac{\partial f_w(x)}{\partial x^T} \right| \cdot \sum_{y=1}^K \left| \dfrac{\partial h_\phi(z; y)}{\partial z^T} \right| p(z_h) p(y).
% \end{align}
% We use a relatively small conditional flow $h_\phi$ to further reduce the computational cost of the summation.
% In our experiments, the conditional flow is approximately 8 times faster then the unconditional one.
Note, that we pass only the deepest components $z_f$ to the conditional flow $h_\phi$, instead of the hole vector $[z_f, z_\mathrm{aux}]$.
In our experiments (Section~\ref{sec:exp-dim}) we found it to be an essential part of our model.
% \textcolor{red}{We pull the first Jacobian determinant out of the sum \eqref{eq:mll} as it is independent of $y$ and can pass $x$ trough $f_\theta$ only once.}

\subsection{Learning of Semi-Conditional Normalizing Flows}
\label{sec:SCNF-opt}
The parameters $\theta = \{w, \phi\} $ of the model \eqref{eq:scnf} are estimated via maximum likelihood approach \eqref{eq:ssl-objective}.
Normalizing flows provide us with tractable log-likelihood function $\log p_\theta(x, y)$ \eqref{eq:scnf} along with marginal log-likelihood $\log p_\theta(x)$.
Therefore, we can compute a gradient $\nabla_\theta \mathcal{L}(\theta)$ of the objective \eqref{eq:ssl-objective} and use a stochastic gradient optimization.
\begin{align}
\label{eq:scnf}
   &\log p_\theta(x, y) =\log \left| \tfrac{\partial f_w(x)}{\partial x^T} \right| + \log \left| \tfrac{\partial  h_\phi(z_f; y)}{\partial z_f^T} \right| \\
    \nonumber&+ \log \N(z_h|0, I) + \log \N(z_{\mathrm{aux}}|0, I) + \log p(y)
\end{align}
\textbf{Connection to NF with Learnable Prior.} 
We can treat the second normalizing flow $h_\phi$ as a conditional prior distribution for the first unconditional flow $f_w$.
% A simple example of such a learnable prior is a Gaussian mixture model (GMM), where each component is a Gaussian distribution $\mathcal{N}(z_f \vert \mu_y, \Sigma_y)$.
A simple example of such a learnable prior is a Gaussian mixture model (GMM) where each of mixture components is a Gaussian distribution $\mathcal{N}(z_f \vert \mu_y, \Sigma_y)$ corresponded to one of the class label $y$.
We compare a performance of the proposed conditional normalizing flow and the Gaussian model as the conditional part (Section~\ref{sec:exps}).
A gold standard to find parameters of GMM is an expectation maximization algorithm \cite{mackay2003information}.
We also adapt this algorithm for our model (see Appendix~\ref{sec:app-em}) and compare it with direct optimization (Section~\ref{sec:mnist}).
We, however, did not find a significant difference between them. 

\vspace{-10pt}

\section{Experiments}
The PyTorch \cite{paszke2019pytorch} implementation is available at GitHub\footnote{\url{https://github.com/bayesgroup/semi-supervised-NFs}}.
\label{sec:exps}
\subsection{Toy Semi-Supervised Classification}
\label{sec:exp-toy}
\begin{table}[]
\begin{adjustbox}{width=\columnwidth,center}
\begin{tabular}{cccccc}
\toprule
\multirow{2}{*}{$f_w$} & \multirow{2}{*}{$h_\phi$} & \multicolumn{2}{c}{Moons} & \multicolumn{2}{c}{Circles} \\
                            &                           & Error,\%       &   NLL              & Error,\%         & NLL       \\ \midrule
\multirow{2}{*}{GLOW}       & GLOW                      &  0.6 $\pm$ 0.4 &   1.11 $\pm$ 0.02  &  5.0 $\pm$ 1.9   &   1.68 $\pm$ 0.13      \\
                            & GMM                       &  1.2 $\pm$ 1.2 &   1.15 $\pm$ 0.01  &  14.2 $\pm$ 9.0  &   1.28 $\pm$ 0.15      \\ \midrule
\multirow{2}{*}{FFJORD}     & GLOW                      &  0.3 $\pm$ 0.2 &   1.12 $\pm$ 0.03  &  6.2 $\pm$ 2.2   &   1.7 $\pm$ 0.2         \\
                            & GMM                       &  5.3 $\pm$ 6.3 &   1.15 $\pm$ 0.06  &  25 $\pm$ 2      &   0.97 $\pm$ 0.05        \\ \bottomrule
\end{tabular}
\end{adjustbox}
\caption{
Test accuracy and negative log-likelihood (NLL) for different models on two toy datasets (see visualization at Appendix.~\ref{sec:app-toy}).
$f_\theta$ corresponds to the unconditional part and $h_\phi$ to the conditional part of SCNF model (Fig.~\ref{fig:scnf-model}). 
GMM stands for Gaussian mixture model, and Glow is a flow-based model.
Glow conditional flow $h_\phi$ in terms of test error outperforms GMM for different types of unconditional flow on both datasets. 
Taking into account values of uncertainties, we see that unconditional FFJORD gives roughly the same test error and negative log-likelihood as unconditional Glow.
% \vspace{-0.5em}
}
\label{tab:toy}
\end{table}
We train proposed Semi-Conditional Normalizing Flow on toy 2-dimensional problems: moons and concentric circles. 
For each problem the training dataset consists of 1000 objects and only 10 of them are labeled.
We consider Gaussian mixture model and Glow as the conditional flow $h_\phi$.
For the unconditional flow $f_\theta$ we take Glow and recently proposed FFJORD \cite{grathwohl2018ffjord} models.
We do not use multiscale architecture with dimension reduction as we have only 2-dimensional input.
% A multilayer Glow model has more parameters than GMM.
% We, therefore, use smaller unconditional part when use Glow then one when use GMM as a conditional part (1 and 4 layers respectively) to make the number of parameters roughly the same.
% We use 1 and 4 layers for the unconditional model when use Glow and GMM for a conditional part, respectively, since Glow has more parameters.
% To make the comparison more fair we use large unconditional flow when GMM as a conditional part and smaller when use Glow, since the latter has more parameters. 
We observe that models with Glow conditioning archive lower test error in comparison with Gaussian Mixture Model conditioning.
To make the fair comparison, we use roughly the same number of parameters for each SCNF model.
Quantitative results can be seen at Tab.~\ref{tab:toy} and visualization at Appendix.~\ref{sec:app-toy}.
% \textcolor{orange}{We observe that Glow + Glow setup (unconditioned flow $f_w$ is Glow and conditioned flow $h_\phi$ is Glow too) shows the least test error on both datasets}

\subsection{Semi-Supervised Classification on MNIST}
\begin{table}[]
    \begin{adjustbox}{width=\columnwidth,center}
    \begin{tabular}{@{}ccccc@{}}
    \toprule
    Model                      & Optimisation            & ${L}_{\mathrm{clf}}$ & Error, \% & Bits/dim \\ \midrule 
    \citet{kingma2014semi}   & VI                      & \cmark                  &    3.3 $\pm$ 0.1      &  -        \\ \midrule
    \multirow{3}{*}{\begin{tabular}[c]{@{}c@{}}SCNF-GLOW\\ \textcolor{gray}{(Ours)}\end{tabular}}  & SGD                     & \xmark                  &    1.9 $\pm$ 0.3      &    1.145 $\pm$ 0.004      \\ \cmidrule(l){2-5}
                              & \multirow{2}{*}{EM-SGD} & \cmark                  &    2.0 $\pm$ 0.1       &    1.151 $\pm$ 0.010      \\ 
                              &                         & \xmark                  &    {1.9 $\pm$ 0.0}      &    1.146 $\pm$ 0.002      \\ \midrule 
    \multirow{3}{*}{\begin{tabular}[c]{@{}c@{}}SCNF-GMM\\ \textcolor{gray}{(Ours)}\end{tabular}} & SGD                     & \xmark                  &    14.2 $\pm$ 2.4      & 1.143 $\pm$ 0.011     \\ \cmidrule(l){2-5}
                              & \multirow{2}{*}{EM-SGD} & \cmark                  &    16.9 $\pm$ 5.3      & {1.141 $\pm$ 0.006}      \\
                              &                         & \xmark                  &    13.4 $\pm$ 2.8      &   1.145 $\pm$ 0.005     \\ \bottomrule
    \end{tabular}
    \end{adjustbox}
    \caption{Test error and bits per dimension on MNIST dataset (lower better). 
    We use 100 labelled objects to train the models (averaging done over 3 different splits). 
    % SCNF stands for the proposed Semi-Conditional Normalizing Flows with the same unconditional flow and different conditional parts (see Fig.~\ref{fig:scnf-model}).
    SCNF stands for Semi-Conditional Normalizing Flows with the same unconditional flow and different conditional parts.
    % SGD is a direct gradient optimisation of the objective, EM-SGD is an expectation maximisation algorithm with one gradient step on M-step and VI means variational inference.
    SGD is a direct gradient optimisation of the objective, EM-SGD is an expectation maximisation algorithm, and VI is a variational inference.
    $L_\mathrm{clf}$ is an additional classification loss.
    We found that the proposed SCNF-GLOW model outperforms VAE-based approach \cite{kingma2014semi}.
    % The model with GMM as the conditional part, however, achieves the best bits/dim score.
    % We also did not find significance difference of SGD and EM-SGD methods.
    }
    \label{tab:mnist-acc}
    % \vspace{-1.2em}
\end{table}
\label{sec:mnist}
We demonstrate performance of the proposed model on MNIST dataset \cite{lecun1998gradient}.
% To model semi-supervised setting we used the standard protocol \cite{kingma2014semi}.
The standard protocol \cite{kingma2014semi} was used to model semi-supervised setting.
We split the training set of size 60,000 into labelled and unlabelled parts.
The size of labelled part equals to 100.
% We perform the splitting a number of times and average the performance.
% We average the performance of the algorithm on 3 different random splits.
The algorithm performance was averaged on 3 different random splits of the dataset.
% We use Glow-like architecture \cite{kingma2018glow}, but with additional checkerboard masks along with channel ones for unconditional flow $f_w$ (see Appendix~\ref{sec:app-arch} for details).
We use Glow architecture \cite{kingma2018glow} for  an unconditional part (see Appendix~\ref{sec:app-arch}).
% The dimension of the deepest component $z_f$ is equal to 196 (4 times reduction).
% We reduce the dimension of an input 4 times from 784 to 196 with multi-scale architecture.
% The dimension 784 of an input $x$ was reduced in 4 times to 196 of $z_f$ with a multi-scale architecture of the unconditioned flow $f_w$.
% The multi-scale architecture of the unconditioned flow $f_w$ reduces input size 4 times, from 784 to 196.
We reduce the size of an input with a multi-scale architecture from 784 for $x$ to 196 for $z_f$.
For a conditional part $h_\phi$ we compare Glow architecture with conditional affine coupling layers (SCNF-GLOW) and Gaussian mixture model (SCNF-GMM).
We compare our Semi-Conditional Normalizing Flow (SCNF) model with VAE-based approach \cite{kingma2014semi}, which also uses architecture with dimension reduction.

The results can be seen at Tab.~\ref{tab:mnist-acc}.
We observe that the proposed SCNF model with Glow-based conditional part outperforms VAE-based model.
The GMM conditioning seems to be not sufficiently expressive for this problem and shows much poorer performance.
% However, SCNF-GMM model achieves the best bits/dim score.
We also did not find any difference between the performance of SGD and EM-SGD optimization algorithms.
% Samples from the generative models can bee seen at Appendix~\ref{sec:mnist-samples}.

% To examine an underling dependencies in the model we plot reconstructions from the same latent representation with different class labels at Fig.~\ref{fig:rec}.
% We observe, that latent representations define a style and fine-granted features while $y$ defines only the digit to reconstruct.

% \subsection{Dimension Reduction}
% \label{sec:exp-dim}
% % We examine an impact of the dimension reduction technique described in Section~\ref{sec:SCNF-model}.
% An impact of the dimension reduction technique was examined (see Section~\ref{sec:SCNF-model}).
% We use the SCNF-GLOW architecture from the previous experiment (Section~\ref{sec:mnist}) and vary a dimension of the latent representation $z_f$.
% Notable that with the dimension growth the model is prone to overfit, and it results in nearly random guess test performance.
% For more detailed results see Appendix~\ref{sec:app-dim}.

\subsection{Dimension Reduction}
\label{sec:exp-dim}

\begin{figure}[t]
    \centering
    \includegraphics[width=1.0\columnwidth]{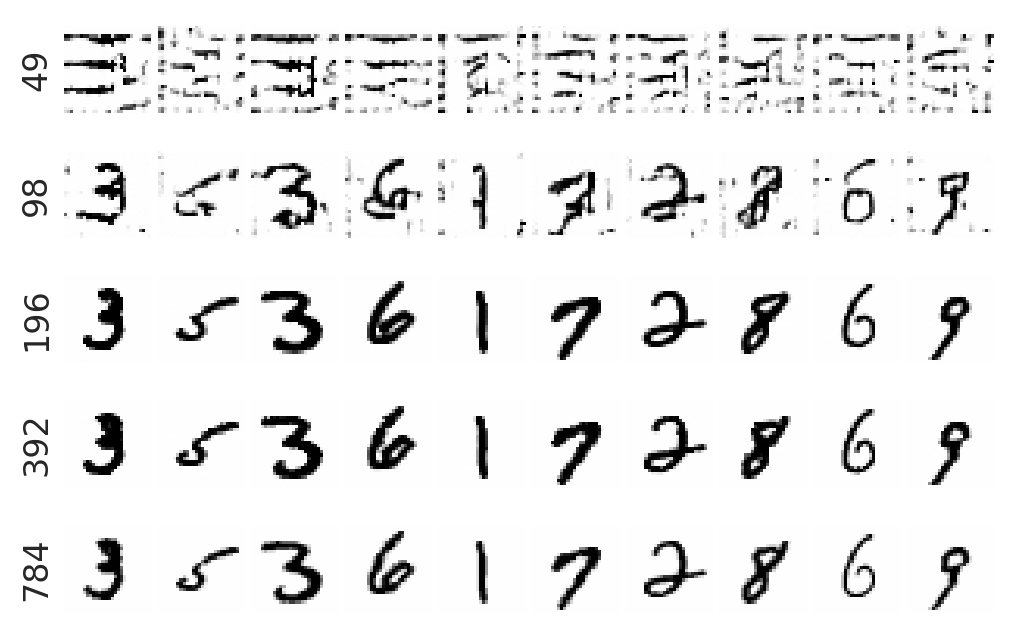}
    \caption{
    Reconstructions of images using different number of deepest hidden components $z_f$. 
    From top to bottom: 49, 98, 196, 392, 784.
    The latter one corresponds to real images as this is the hole representation and the architecture is invertible.
    We zeroed the auxiliary components $z_\mathrm{aux}$ when perform the reconstruction.
    We found that the deepest 196 components provide quite accurate reconstructions.
    }
    \label{fig:mnist-rec}
\end{figure}
\begin{table}[t]
    \centering
    \begin{adjustbox}{width=0.8\columnwidth,center}
    \begin{tabular}{ccc}
        Dimension & Test Error, \% & Train Error, \% \\
        \toprule
        48 & 2.6 & 0\\
        98 & 2.0 & 0\\
        196 & \textbf{1.9} & 0\\
        392 & 61.4 & 0\\
        784 & 91.1 & 0\\
        \hline
    \end{tabular}
    \end{adjustbox}
    \caption{Test and train erros on MNIST dataset for different dimensions the deepest components $z_f$ that we pass to conditional model $h_\phi(z_f; y)$ (see Fig.\ref{fig:scnf-model}).}
    \label{tab:dim}
\end{table}

% \citet{dinh2016density} proposed a multi-scale architecture for normalizing flows and showed that the deepest components of latent representation contains almost all semantic information about an input object.
% We perform the same experiment and try to reconstruct an input object using different number of deepest components of its latent representation $z_f$, zeroing the other components $z_\mathrm{aux}$.
% The results can be seen at Fig.~\ref{fig:mnist-rec}.

Training classification model in a high-dimensional space with only a few labelled examples may lead to overfitting.
In \citet{kingma2014semi} authors showed that semi-supervised classification methods benefits from using low-dimensional representation of objects.
In Section~\ref{sec:dim} we proposed a natural dimension reduction technique for our model.
In this Section we examined an impact of the proposed technique.

We use the SCNF-GLOW architecture from the previous experiment (Section~\ref{sec:mnist}) and vary a dimension of the latent representation $z_f$.
The results can bee seen at Tab.~\ref{tab:dim}.
We find that with the dimension growth the model is prone to overfitting and results in nearly random guess test performance.
To demonstrate an information that remains in the latent representation we reconstruct an image using $z_f$ and zeroing the auxiliary components $z_{\mathrm{aux}}$.
The corresponding reconstructions can be found at Fig.~\ref{fig:mnist-rec}.
% We observe that 196 deepest components provide us with quite precise reconstructions.
Surprisingly, reconstructions from 49 deepest components do not look like original images, while conditional flow is still able to achieve low test error on test set.

% For more detailed results see Appendix~\ref{sec:app-dim}.

% In our experiments for MNIST dataset the dimension of a raw data equals 784.
% We examine different sizes of low-dimension representation $z_f$ we pass to a conditional part $h_\phi(z_f; y)$.
% and the number of labelled objects is equal to 100.
% We tested different sizes of representation we pass to a conditional part of our architecture.
% Performance of final models can be found at Tab.~\ref{tab:dim}.
% We observe that with growth of the latent representation dimension the accuracy of a classifier significantly drops.

\subsection{Semi-Supervised Data Obfuscation}

\begin{figure}
\centering     %%% not \center
\subfigure[t-SNE Embeddings of $z_f$]{\label{fig:a}\includegraphics[width=0.47\columnwidth]{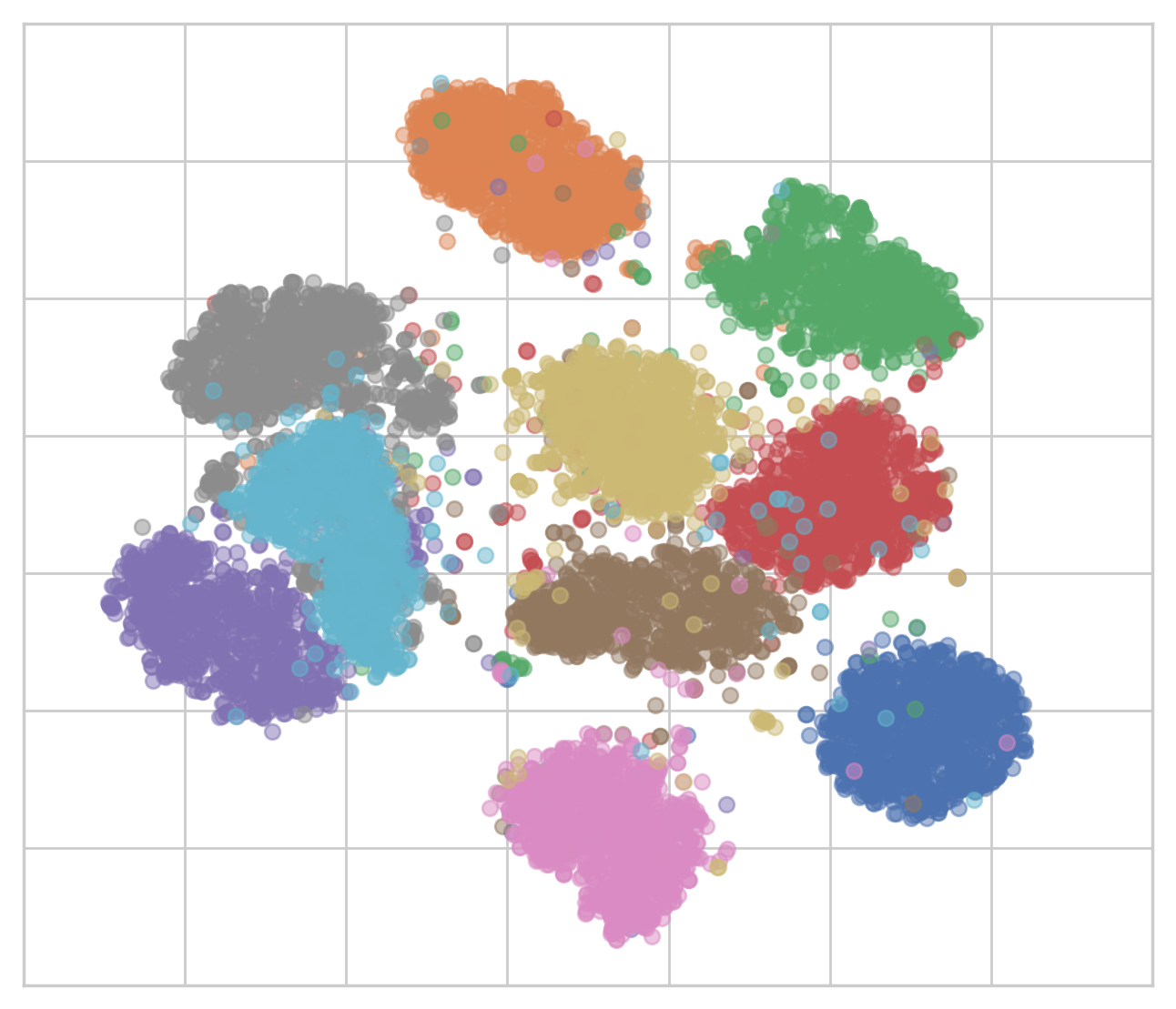}}
~~~
\subfigure[t-SNE Embeddings of $z_h$]{\label{fig:b}\includegraphics[width=0.47\columnwidth]{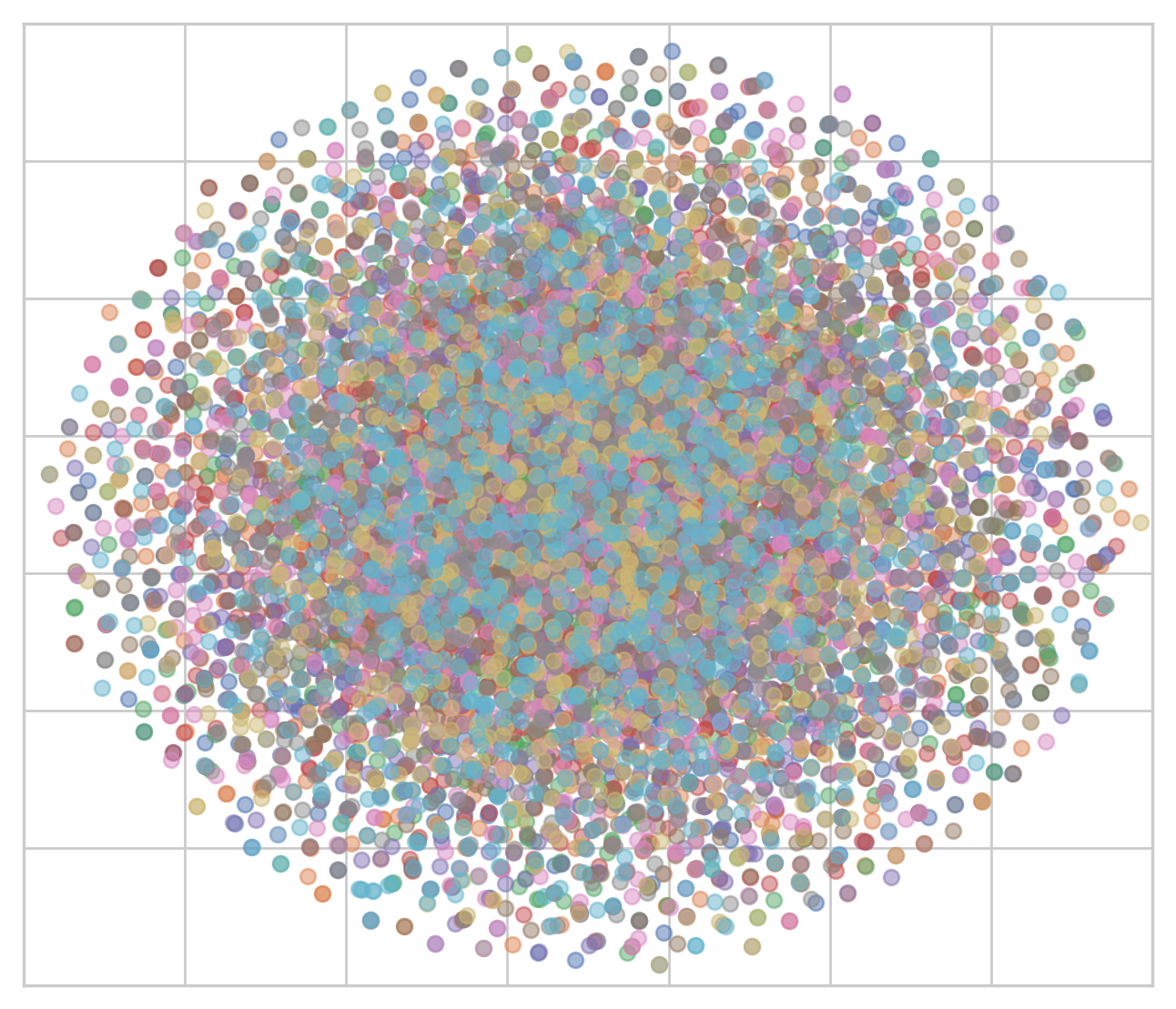}}
\vspace{-0.8em}
\caption{
We embed low-dimension representations of MNIST digits into two dimensional space using t-SNE method.
$z_f$ are from an unconditional part $f_\theta$ and $z_h$ are from a conditional part $h_\phi$.
Different colors correspond to different classes.
We found that embeddings $z_f$ from the unconditional model form clusters for different classes, and embeddings $z_h$ seem to be independent of class variable.\vspace{0.5em}}
\label{fig:2d-embed}
\end{figure}

\label{sec:exp-obf}
% Let us consider a problem of removing an information from an object description.
% For example, it can by discriminatory information about gender and race of a person photo.
\label{dataobf}
Modern machine learning algorithms have become widely used for decisions that impact real lives.
% These algorithms, however, tend to be biased -- relies on features e.g., gender and race, which may not have casual relations to target variable, but introduce discrimination.
These algorithm have to be non-discriminative and does not rely on such features as gender and race.
One way to reduce the discrimination is to construct object representations, that are independent of these discrimination-inducing features.
% In our work, we found that latent representations $z_h$ produced by a conditional model $h_\phi$ are not dependent of class variable $y$, while contain all other semantic information about object $x$. In the next paragraph we assume that $z_f=f_w(x)$ contains all semantic information about object $x$, see Section~\ref{sec:exp-dim}.
% In our work, we found that latent representations $z_h$ produced by a conditional model $h_\phi$ are not dependent of class variable $y$, while contain all other semantic information about object $x$. 
% To see it, let us assume that $z_f=f_w(x)$ contains almost all semantic information about object $x$, because $x$ can be quite accurate restored from $z_f$ (when $z_f$ dimension equals to 196, see Figure~\ref{fig:mnist-rec}). 
We consider a class label to be such discriminative information, and focus on finding a representation of $x$ that is independent of $y$.
% We found that $z_h$ from a conditional flow can be such a non-discriminative representation of an object.
The latent representation $z_f$ contains almost all semantic information about an object $x$, as $x$ can be quite accurate restored from $z_f$ when its dimension equals to 196 (see Figure~\ref{fig:mnist-rec}) and we assume that $x \!=\! f^{-1}_w(z_f)$.
However, it means that $z_f$ contains information about $y$ and, therefore, does not suitable.
On the other hand, we found that the representation $z_h$ from a conditional flow is independent of $y$, but still contains all the other information about $z_f$ and therefor $x$, since $z_f \! =\! h_\phi^{-1}(z_h; y)$.
Therefore, $z_h$ can be used as a non-dicriminative representation of an object $x$.

% The intuition, why latent representations $z_h$ are not dependent of class variable $y$, can be explained by properties of normalizing flows. 
% The conditional flow $h_\phi$ is invertible given a class label $y$, i.e., $z_f \!=\! h_\phi^{-1}(z_h; y)$.
% Therefore, latent representation $z_h$ contains all information about $z_f$ except perhaps for a class label $y$.
% Although $z_h$ may contain information about class label $y$, it contradicts with log-likelihood \eqref{eq:scnf} maximization.
The intuition, why latent representations $z_h$ are not dependent of class variable $y$, can be explained by contradiction with log-likelihood \eqref{eq:scnf} maximization.
The likelihood, as a function of $h_\phi$ consists of two terms: a determinant of Jacobian ${\partial h_\phi(z_f; y)}/{\partial z_f^T}$ and a prior density $\mathcal{N}\left(z_{h} \vert 0, I\right)$.
The Jacobian term encourage the transformation $h_\phi$ to increase a volume, while the prior term prevent it from covering the whole space.
The conditional flow $h_\phi$ models different transformations for each class label $y$. 
% Therefore, it is able to share the same high-density support in latent space for objects from different classes and increase the determinant of Jacobian while preserve the prior density the same.
Therefore, it is able to fill the same high-density area in the latent space for objects from different classes and increase the determinant of Jacobian while preserving the prior density the same.
This leads to a higher likelihood and reduces the information about class label $y$.

In order to provide an evidence for this intuition, we trained two fully-connected classifiers of the same size in a space of $z_f$ and $z_h$.
For both classifiers we were able to fit training data perfectly. 
However, the classifier that was trained with representations $z_h$ has nearly random accuracy on test data, while the other one achieves almost perfect performance.
To illustrate that $z_f$ depends on $y$ and $z_h$ does not, we embed these latent representations using t-SNE \cite{maaten2008visualizing} into 2-dimensional space (Fig.~\ref{fig:2d-embed}).
In Appendix~\ref{sec:mnist-cond-rec} we also explore what type of information contains in $z_h$.

% To verify this intuition we train two classifiers $C_f$ and $C_h$ to classify images from MNIST using their latent representations $z_f$ and $z_h$ respectively.
% We found that both classifiers fit training data ideally, however, test accuracies differ a lot: 98\% for $C_f$ and 10\% for $C_h$, i.e. $C_h$ fails to find any reasonable dependency between $z_h$ and $y$.

% We can use this property for data obfuscation [], where we want to exclude some specific information from an object description $x$.
% % Indeed, the representation $z_h$ still contains information about the object $x$ as we can ideally reconstruct it using only additional information about the class.
% Indeed, we can ideally reconstruct $x$ from $z_h$ using only additional information about the class.
% Thus, we exclude the information about the class variable $y$, wile save all the other information about the object $x$.
% If we want our model to not use information about some property $y$ we can use embeddings $z_h$ instead raw data.

\section{Conclusion \& Discussion}
This work investigates the idea of using deep generative models for semi-supervised learning, as well as, expands an area of normalizing flows applications.
We propose \textit{Semi-Conditional Normalizing Flows} -- a new architecture for semi-supervised learning that provides a computationally efficient way to learn from unlabeled data.
% \textcolor{gray}{Dimensional reduction -- one of the key ingredients of the model -- however, is not forced directly. 
% Even though we always observe it in practice, this drawback may affect the quality of the model and may need an additional investigation.}
Although the model does not directly force the low-dimensional representations to be semantically meaningful, we always observe it in practice.
This effect is crucial for our model and needs an additional investigation. 

In experiments, we show that the model outperforms semi-supervised variational auto-encoders \cite{kingma2014semi}. 
We refer the improvement to end-to-end learning and a tractable likelihood, which optimization does not require to use approximate variational methods.
We also show, that the proposed model has a notable effect of data obfuscation (Section \ref{dataobf}), that is provided by the model design and properties of normalizing flows.
This effect can be also used for semi-supervised domain adaptation as has been proposed recently by \cite{ilse2019diva}.

On more complex datasets the flows tend to ignore a class label $y$, that in case of Gaussian Mixture prior is caused by similar mixture components for different classes.  
As proposed in Appendix D of \cite{kingma2018glow}, the effect can be cured by using classification loss on the hidden components of the flow.
We found that classification loss improves the performance of our model on more complex datasets, however, recent methods based on data augmentation  \cite{berthelot2019mixmatch, QizheXie} tend to show a better performance. 

A concurrent work by \citet{izmailov2019semi} proposes a model for semi-supervised learning based on normalizing flows and GMM prior.
The aforementioned model provides a superior performance comparing to our method and proven to scale to larger datasets.
This is likely due to \emph{i)} using consistency regularization \emph{ii)} using fixed parameters of Gaussian mixture and \emph{iii)} difference of architectures. 
The paper also provides a wide empirical evaluation, we recommend to refer it for more results on semi-supervised learning with normalizing flows.

\section*{Acknowledgements}
This research is in part based on the work supported by Samsung Research, Samsung Electronics. This research was supported in part through computational resources of HPC facilities at NRU HSE

\bibliography{example_paper}
\bibliographystyle{icml2019}

\appendix

{
}
\newpage

\section{Conditional Samples from Semi-Conditional Normalizing Flows}
\label{sec:mnist-cond-rec}
\begin{figure}
    \centering
    \includegraphics[width=1\columnwidth]{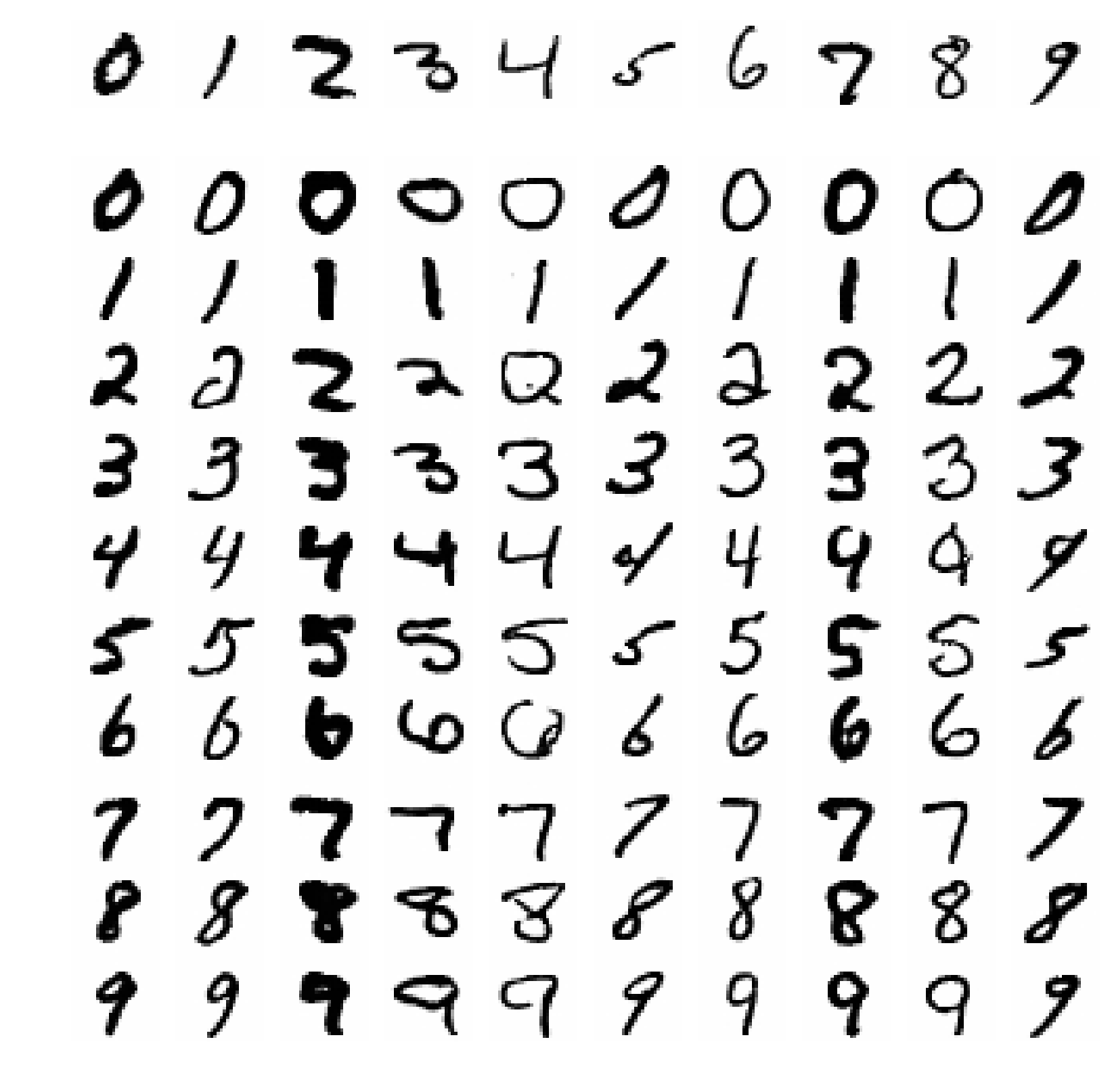}
    \caption{
    Reconstructions from the SCNF-GLOW model with different labels.
    Top row corresponds to source images, that are encoded into latent space with the conditional flow using true labels.
    Then, we reconstruct this latent representations using different class-labels.
    We can see that the latent representation define a style but not a particular class.
    % We can use this to transfer style from one digit to another.
    }
    \label{fig:mnist-cond-rec}
\end{figure}
The proposed SCNF model, as have been argued in Section~\ref{sec:exp-obf}, removes the information about a class label $y$ from the latent representation $z_h$.
To demonstrate what type of information contains in $z_h$, we choose 10 different digit images and encode them with their true class labels.
Then, for each image we fix its latent representation $z_h$ and decode it using different class labels $y$.
We zeroed $z_\mathrm{aux}$ to not use information from it.
The results can be seen at Fig.~\ref{fig:mnist-cond-rec}.
We found that $z_h$ seems to contain information about form and style of a digit.

\section{Expectation Maximization algorithm for Semi-Conditional Normalizing Flow}
\label{sec:app-em}
Parameters of the proposed model are found via maximum likelihood approach (Section \ref{sec:ssl}, \ref{sec:SCNF-opt}).
For labelled data we directly maximize a joint log-likelihood $\log p_\theta(x, y)$ \eqref{eq:scnf}.
For unlabelled objects we have to maximize a marginal log-likelihood \eqref{eq:scnf-mll}:
\begin{equation*}
    \log p_\theta(x) = \log \sum_{y=1}^K p_\theta(x \vert y) p(y).
\end{equation*}
It can be seen as a mixture model with components $p_\theta(x|y)$ for different $y$.
The class variable $y$ in this case plays a role of a hidden variable.
A gold standard for training models with hidden variables is an Expectation Maximization algorithm.

Let us introduce an auxiliary distribution $q(y)$ over the latent variable $y$ and define the following decomposition of the marginal log-likelihood:
\begin{equation*}
    \log p_\theta(x) = \underbrace{\mathbb{E}_{q(y)} \log p_\theta(x, y)}_{\mathcal{L}(q, \theta)} + \mathrm{KL}(q(y) \Vert p_\theta(y \vert x)),
\end{equation*}
where the second term is a Kullback-Leibler divergence between true posterior over $y$ and its approximation $q(y)$.
% Since the divergence is non-negative the first term $\mathcal{L}(q, \theta)$ is a lower bound for $\log p_\theta(x)$.
The Expectation Maximization algorithm optimizes the marginal log-likelihood with an iterative process of two subsequent steps:
\begin{align*}
    \text{E-step:} & ~~~~~~ q^*(y) = \arg\min_q ~ \mathrm{KL}(q(y) \Vert p_{\theta_{old}}(y \vert x)) \\
    \text{M-step:} & ~~~~~~ \theta_{new} = \arg\max_\theta ~ \mathcal{L}(q^*, \theta).
\end{align*}

The first step can be done analytically since the KL-divergence is non-negative and equals to 0 iff $q(y) = p_{\theta_{old}}(y|x) = \frac{p_{\theta_{old}}(x, y)}{p_{\theta_{old}}(x)} $.
This step ensures that the lower bound is equal to the true log-likelihood $\mathcal{L}(q^*, \theta_{old}) = \log p_\theta(x)$ for the previous parameters $\theta_{old}$.
Then, during M-step we optimize the lower bound and, therefor, the marginal likelihood.
Since in the proposed model the true posterior is tractable we can use this analytic solution.
The optimization problem on the M-step is intractable, unfortunately.
However, we can use a gradient method for an approximate optimization.
We use one gradient step from previous parameters $\theta_{old}$ on the M-step.
The final EM algorithm for the proposed model looks as follows:
\begin{align*}
    \text{E-step:} & ~~~~~~ q^*(y) = p_{\theta_{old}}(y \vert x) \\
    \text{M-step:} & ~~~~~~ \theta_{new} = \theta_{old} + \nabla_\theta \mathcal{L}(q^*, \theta)\big|_{\theta = \theta_{old}}.
\end{align*}
We refer to this algorithm as EM-SGD.
\section{Conditional FFJORD}
\label{appcfd}
Our model consists of unconditional and conditional flows.
We showed how GLOW \cite{kingma2018glow} can be efficiently conditioned and used it in experiments (Table \ref{tab:toy}). 
Moreover, we propose a natural way to condition FFJORD \cite{grathwohl2018ffjord}. This flow maps objects to latent space by solving the following ordinary differential equation:
$$\frac{dz(t)}{dt} = f_{\theta}(z(t), t),$$
where $z(t)$ is a trajectory that draws latent representation $z(0)$ to objects $x = z(T)$. 
Dynamics function $f$ is a neural network that takes $z(t)$ and $t$ input and returns $\frac{dz(t)}{dt}$. 
We propose to feed also a class label $y$ into $f$ to make conditional FFJORD. So, we consider to solve the following equation:
$$\frac{dz(t)}{dt} = f_{\theta}(z(t), t, y)$$
We assume that conditional FFJORD will perform better than conditional GLOW because as it was shown \cite{grathwohl2018ffjord} FFJORD is more expressive than GLOW. But our current results didn't give significant improvements, so we are going to investigate this problem in the future.

\section{Toy Experiments Visualization}
\label{sec:app-toy}
\begin{figure}[t!]
\centering     %%% not \center
\subfigure[Circles]{\label{fig:a}\includegraphics[width=0.75\columnwidth]{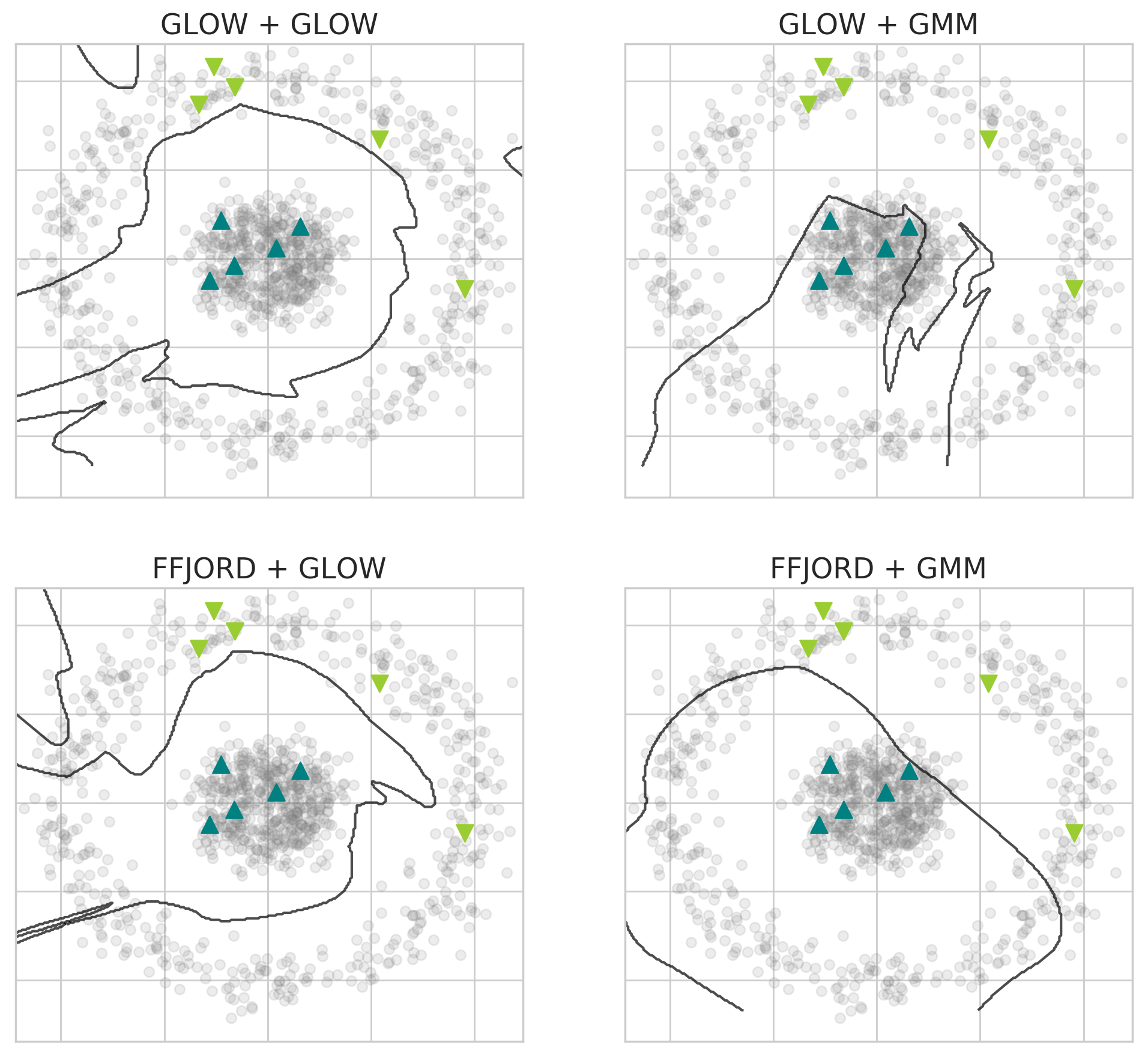}}
\subfigure[Moons]{\label{fig:b}\includegraphics[width=0.75\columnwidth]{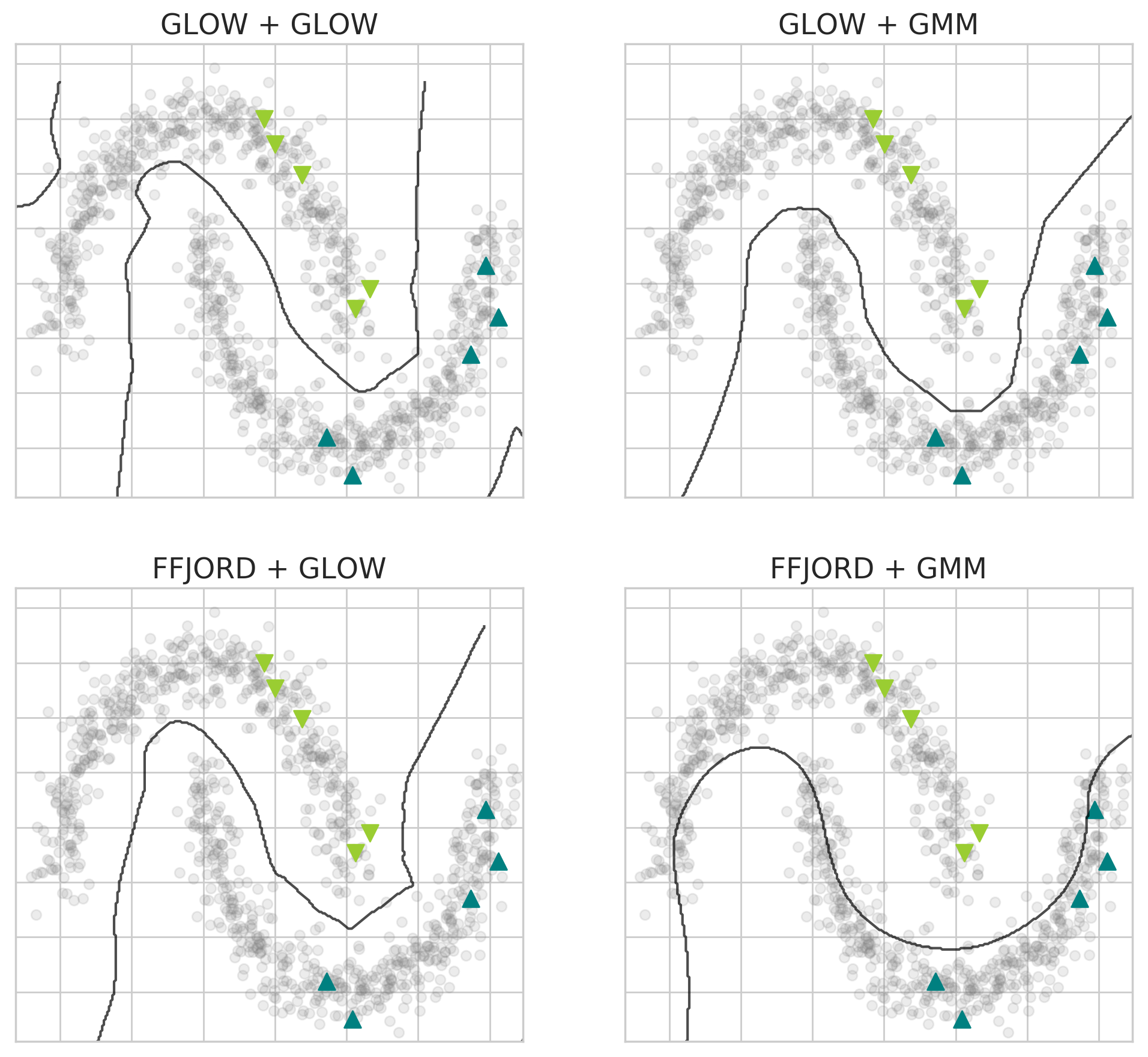}}
\vspace{-0.8em}
\caption{
Visualization of classification curves for different SCNF models on toy datasets.
Gray objects are unlabelled and colored are labelled points.
Black line corresponds to a decision boundary of the corresponding classification rule.
On Circles problem, we found that FFJORD with GMM is prone to overfit labeled data, that leads to poor test accuracy.
We also did not find that regularization helps.
This effect, however, disappears since we use more labeled data.}
\label{fig:toy-clf}
\end{figure}
We plot decision boundaries for models trained on toy datasets (Section~\ref{sec:exp-toy}) at Fig.~\ref{fig:toy-clf}.

\section{Normalizing Flow Architecture for MNIST}
\label{sec:app-arch}
For experiments on MNIST dataset we use Glow-based architecture for both unconditional and conditional parts.
For coupling layers use two types of masks: checkerboard and channel-wise.
We use residual architecture \cite{he2016deep} with 4 residual blocks with hidden size 64 as a neural network in all coupling layers.
\subsection{The Unconditional Flow $f_\theta$}
We also transform pixels to logits as have been proposed in \cite{dinh2016density}.
The sequence of transformations in the unconditional flow $f_\theta$ looks as follows:
\lstset{
   basicstyle=\fontsize{9}{9}\selectfont\ttfamily
}

\begin{lstlisting}
    ToLogits()
    InvertibleConv2d(1)
    MaskedCouplingLayer(mask=checkerboard)
    ActNorm(1)
    InvertibleConv2d(1)
    MaskedCouplingLayer(mask=checkerboard)
    ActNorm(1)
    InvertibleConv2d(1)
    MaskedCouplingLayer(mask=checkerboard)
    ActNorm(1)
    SpaceToDepth(2x2)
    InvertibleConv2d(4)
    CouplingLayer(mask=channel)
    ActNorm(4)
    InvertibleConv2d(4)
    CouplingLayer(mask=channel)
    ActNorm(4)
    FactorOut([4, 14, 14] -> [2, 14, 14])
    InvertibleConv2d(2)
    MaskedCouplingLayer(mask=checkerboard)
    ActNorm(2)
    InvertibleConv2d(2)
    MaskedCouplingLayer(mask=checkerboard)
    ActNorm(2)
    InvertibleConv2d(2)
    MaskedCouplingLayer(mask=checkerboard)
    ActNorm(2)
    SpaceToDepth(2x2)
    InvertibleConv2d(8)
    CouplingLayer(mask=channel)
    ActNorm(8)
    InvertibleConv2d(8)
    CouplingLayer(mask=channel)
    ActNorm(8)
    FactorOut([8, 7, 7] -> [4, 7, 7])
    InvertibleConv2d(4)
    MaskedCouplingLayer(mask=checkerboard)
    ActNorm(4)
    InvertibleConv2d(4)
    MaskedCouplingLayer(mask=checkerboard)
    ActNorm(4)
    InvertibleConv2d(4)
    MaskedCouplingLayer(mask=checkerboard)
    ActNorm(4)
    InvertibleConv2d(4)
    CouplingLayer(mask=channel)
    ActNorm(4)
    InvertibleConv2d(4)
    CouplingLayer(mask=channel)
    ActNorm(4)
\end{lstlisting}
The deepest 196 components of the output of this model then pass to the conditional flow $h_\theta$.

\subsection{The Conditional Flow  $h_\theta$}
The sequence of transformations in the conditional flow looks as follows:
\begin{lstlisting}
    ActNorm(196)
    InvertibleConv2d(196)
    ConditionalCouplingLayer(mask=channel)
    ActNorm(196)
    InvertibleConv2d(196)
    ConditionalCouplingLayer(mask=channel)
    ActNorm(196)
    InvertibleConv2d(196)
    ConditionalCouplingLayer(mask=channel)
    ActNorm(196)
    InvertibleConv2d(196)
    ConditionalCouplingLayer(mask=channel)
    FactorOut([196, 1, 1] -> [98, 1, 1])
    ActNorm(98)
    InvertibleConv2d(98)
    ConditionalCouplingLayer(mask=channel)
    ActNorm(98)
    InvertibleConv2d(98)
    ConditionalCouplingLayer(mask=channel)
    ActNorm(98)
    InvertibleConv2d(98)
    ConditionalCouplingLayer(mask=channel)
    ActNorm(98)
    InvertibleConv2d(98)
    ConditionalCouplingLayer(mask=channel)
\end{lstlisting}
As a prior we use Gaussian distribution with zero mean and learnable diagonal covariance matrix.

\end{document}